\documentclass[letterpaper, 10 pt, conference]{ieeeconf}
\IEEEoverridecommandlockouts                            
\usepackage{hyperref}
\usepackage{amsmath,amssymb,mathtools}
\usepackage{algorithm}
\usepackage{bm}
\usepackage{cite}
\usepackage{graphicx,subfigure}
\usepackage{booktabs}
\usepackage{chemfig}
\usepackage{multirow}
\usepackage{url}
\usepackage{graphicx} 
\usepackage{setspace}
\usepackage{pgfplots}
\usepgfplotslibrary{groupplots}
\pgfplotsset{width=0.9\columnwidth, height=0.62\columnwidth,compat=1.9}

\newcommand\mtlarge{\fontsize{8pt}{10pt}\selectfont}
\newcommand\markSize{1.5}
\newcommand\pointInt{10}
\pgfplotsset{every axis/.append style={
xlabel={$x$},          
ylabel={$y$},          
label style={font=\mtlarge},
tick label style={font=\mtlarge},
legend style={font=\mtlarge}
}}

\usepackage{theorem}
\newtheorem{lemma}{Lemma}
\newtheorem{theorem}{Theorem}
\newtheorem{proposition}{Proposition}
\newtheorem{assump}{Assumption}

\newtheorem{myRemark}{Remark}
\newtheorem{remark}{Remark}

\newcommand{\E}{\mathop{\mathbb{E}}}
\newcommand{\defeq}{\vcentcolon=}

\usepackage{algpseudocode}
\usepackage{xcolor}
\usepackage{amssymb}
\usepackage{pifont}

\allowdisplaybreaks[4]
\newcommand{\be}{\begin{equation}}
\newcommand{\ee}{\end{equation}}

\newcommand{\Hil}{\mathcal{H}}

\newcommand\lf[1]{\underline{f}_{#1}}
\newcommand\uf[1]{\bar{f}_{#1}}

\newcommand\lowg[1]{\underline{g}_{#1}}
\newcommand\ug[1]{\bar{g}_{#1}}

\usepackage{soul}

\ifodd 1
\newcommand{\longversion}[1]{{ #1}} 
\newcommand{\shortversion}[1]{{}} 
\else
\newcommand{\longversion}[1]{{}} 
\newcommand{\shortversion}[1]{{#1}} 
\fi

\ifodd 0
\newcommand\rev[1]{{\color{blue}#1}}
\newcommand{\com}[1]{\textbf{\color{red} (COMMENT: #1)}} 
\else
\newcommand\rev[1]{{#1}}
\newcommand{\com}[1]{}
\fi

\ifodd 0
\newcommand\finalRev[1]{{\color{blue}#1}}
\newcommand{\finalCom}[1]{\textbf{\color{red} (COMMENT: #1)}} 
\else
\newcommand\finalRev[1]{{#1}}
\newcommand{\finalCom}[1]{}
\fi

\title{\LARGE \bf
Primal-Dual Contextual Bayesian Optimization for Control System Online Optimization with Time-Average Constraints}

\author{Wenjie Xu, Yuning Jiang, Bratislav Svetozarevic, Colin N. Jones
\thanks{This work was supported by the Swiss National Science Foundation under the NCCR Automation (grant agreement 51NF40\_180545).}%
\thanks{W. Xu, Y. Jiang, and C. N. Jones are with the Automatic Control Laboratory, EPFL, Switzerland. W. Xu and B. Svetozarevic are with Swiss Federal Laboratories for Materials Science and Technology
(Empa), Switzerland. \tt{\{wenjie.xu, yuning.jiang, colin.jones\}@epfl.ch, bratislav.svetozarevic@empa.ch}}%
}

\begin{document}

\maketitle
\thispagestyle{empty}
\pagestyle{empty}

\begin{abstract}
This paper studies the problem of online performance optimization of constrained closed-loop control systems, where both the objective and the constraints are unknown black-box functions affected by exogenous time-varying contextual disturbances. A primal-dual contextual Bayesian optimization algorithm is proposed that achieves sublinear cumulative regret with respect to the dynamic optimal solution under certain regularity conditions. Furthermore, the algorithm achieves zero time-average constraint violation, ensuring that the average value of the constraint function satisfies the desired constraint. The method is applied to both sampled instances from Gaussian processes and a continuous stirred tank reactor parameter tuning problem; simulation results show that the method simultaneously provides close-to-optimal performance and maintains constraint feasibility on average. This contrasts current state-of-the-art methods, which either suffer from large cumulative regret or severe constraint violations for the case studies presented.
\end{abstract}

\section{Introduction}

Constrained closed-loop control systems are \rev{critical to a wide range of} applications, and optimizing their online performance \rev{by tuning control parameters} is a common challenge. This paper \rev{studies} the time-varying black-box optimization problems that arise in optimizing system performance \rev{with constraints}, where both \rev{the} objective and \rev{the} constraints are unknown. For instance, consider building control, which aims to minimize energy consumption while meeting occupant comfort requirements~\cite{xu2022vabo}. Solving the controller tuning problem in such scenarios is challenging for several reasons. Firstly, it is hard to model the mapping from control parameters to the performance metric of the closed-loop system. Secondly, \rev{unknown constraint violations need to be taken care of} during the optimization process. Finally, \rev{evaluating} the system's performance is often expensive in practice.

To tackle these challenges, Bayesian optimization has shown promise as a sample-efficient, derivative-free black-box optimization method~\cite{xu2022lower}. Bayesian optimization constructs a surrogate model using Gaussian process regression~\cite{williams2006gaussian} and uses this model to guide the sampling of the black-box functions. Variants of Bayesian optimization methods have been proposed to handle constrained optimization problems. One common method is to maximize constrained expected improvement~(CEI)~\cite{gelbart2014bayesian, gardner2014bayesian} to select the next sample in each step. Another line of research developed safe Bayesian optimization~(\rev{Safe BO}) methods by restricting the sampling to only feasible points~\cite{sui2015safe,sui2018stagewise,bergmann2020safe,baumann2021gosafe}.~\rev{These methods have found wide applications in control system optimization. For example, CEI method is applied to trajectory optimization for path following control~\cite{rupenyan2021performance}. Safe BO is applied to tune the PI controller of a room heating system subject to a group of safe constraints~\cite{fiducioso2019safe}.} However, the CEI method may not guarantee constraint feasibility and can suffer from severe constraint violations, while safe Bayesian optimization methods can be too cautious and lose performance due to the strict requirement of sampling feasible points.  

In the context of controller tuning, one additional challenge that needs to be considered is that the unknown objective and constraints are time-varying~\cite{brunzema2022controller}. To address this \rev{challenge}, \rev{several} time-varying variants of Bayesian optimization have been proposed in the literature~\cite{bogunovic2016time,brunzema2022controller,deng2022weighted}. In practice, these variations can be due to the change of \rev{`contextual variables'} observed by the decision-maker before selecting a new set of parameters~\cite{xu2023violation}. \rev{Affected by the time-varying contextual disturbances, guaranteeing constraint satisfaction at every time instant can be challenging and may not be necessary\rev{~\cite{fleming2017time}}. But rather, it can be more of interest to satisfy the \emph{time-average} constraint\rev{~\cite{korda2014stochastic}}, especially when it represents some economic cost that accumulates over time. Examples include performance optimization of a machine subject to fatigue constraints (see, e.g.,~\cite{cannon2009probabilistic}) and data center cooling subject
to the constraints on the number of delayed queries per unit
of time~\cite{kim2012free}.} 

Motivated by the aforementioned observations, \rev{we formulate the time-varying black-box optimization problem as a constrained contextual Bayesian optimization problem}. To \rev{solve it}, we extend the primal-dual Bayesian optimization framework~\cite{zhou2022kernelized} to the contextual setting~\cite{krause2011contextual}. In the proposed algorithm, the contextual variable that impacts the response surface of the black-box functions can be observed at each step \rev{before selecting a new set of parameters}, and then the system performance is optimized over the current context. The detailed contribution is summarized as follows:
\begin{itemize}
    \item We propose a \rev{\textbf{P}rimal-\textbf{D}ual \textbf{C}ontextual \textbf{B}ayesian \textbf{O}ptimization~(PDCBO)} algorithm for time-varying constrained black-box optimization problems. In contrast to the regret with respect to a static optimal value in the non-contextual setting~\cite{zhou2022kernelized}, we provide bounds on the contextual regret \rev{with respect to} the time-varying optimal solutions. Furthermore, our algorithm can achieve \emph{zero} time-averaged constraint violations, even under the \emph{adversarial} time-varying contextual setting;
    
    \item The proposed method is deployed to both sampled instances from Gaussian processes and to a continuous stirred-tank reactor \rev{parameter} tuning problem. Compared to other state-of-the-art methods, our method can simultaneously achieve the lowest cumulative objective while satisfying all the constraints on average.   
\end{itemize}

\section{Problem Formulation}
This work aims to sequentially optimize the control \rev{parameters $\theta_t\in\Theta\subset\mathbb{R}^{n_\theta}$ after observing some contextual variable $z_t\in\mathcal{Z}\subset\mathbb{R}^{n_z}$ in each step $t$, where $\Theta$ is the candidate set of the control parameters and $\mathcal{Z}$ is the set of possible contextual variables}. Our problem can be thus, formulated as follows:
\begin{subequations}
\label{eqn:prob_form}
\begin{align}
\min_{\theta\in\Theta}  \quad& f(\theta, z),\label{eqn:obj} \\
\text{subject to:} \quad & g_i(\theta, z)\leq 0,\quad \forall i\in[N],\label{eqn:constraint}
\end{align}
\end{subequations}
where $[N]$ is defined as the set $\{1, 2, \cdots, N\}\subset \mathbb Z$, $f:\mathbb{R}^{n_\theta}\times\mathbb{R}^{n_z}\to \mathbb{R}$ is the black-box objective function, and $g_i:\mathbb{R}^{n_\theta}\times\mathbb{R}^{n_z}\to\mathbb{R}, i\in[N]$ is a set of black-box constraints to be satisfied.  
We use $\theta^*(z)$ to denote the optimal solution of the problem~\eqref{eqn:prob_form}, which depends on the current context variable $z$. We use $g$ to denote the concatenation $(g_i)_{i=1}^N$ and $g(\theta, z)$ to denote the concatenation $(g_i(\theta, z))_{i=1}^N$.

We make some assumptions regarding the regularities of the problem.
\begin{assump}
\label{assump:support_set}
Both $\mathcal{Z}$ and $\Theta$ are compact.
\end{assump}
Assumption~\ref{assump:support_set} is commonly seen in practice. For example, we can restrict the control parameter set $\Theta$ to a hyper box and often know \rev{an apriori} bound on the range of contextual variables.   
\begin{assump}
\label{assump:bounded_norm} 
$f\in\Hil_{0}, g_i\in\Hil_{i}$, where $\Hil_i,i\in\{0\}\cup[N]$ is a reproducing kernel Hilbert space~(RKHS) equipped with kernel function $k_i(\cdot, \cdot)$~(See~\cite{scholkopf2001generalized}.). Furthermore, $\|f\|\leq C_0, \|g_i\|\leq C_i,\forall i\in[N]$, where $\|\cdot\|$ is the norm induced by the inner product of the corresponding RKHS without further notice. 
\end{assump}
Assumption~\ref{assump:bounded_norm} requires that the underlying functions are regular in the sense of having a bounded norm in an RKHS. In the existing literature, having a bounded norm in an RKHS is a commonly adopted assumption~(e.g.,~\cite{srinivas2012information,zhou2022kernelized}). \rev{Intuitively, it means the black-box function has \rev{a} `smoothness' property at least to a certain level~(See~\cite{scholkopf2001generalized}).} 

\begin{assump}
\label{assump:obs_model}
We only have access to a noisy zero-order oracle, which means each round of query $\theta_t$ with contextual variable $z_t$ returns the noisy function evaluations, 
\begin{subequations}
\begin{align}
y_{0,t}&=f(x_t)+\nu_{0,t}\enspace,\\
y_{i,t}&=g_i(x_t)+\nu_{i,t}\enspace,\quad i\in[N],
\end{align}
\end{subequations}
where $x_t$ denotes the concatenation $(\theta_t, z_t)$ and $\nu_{i,t},i\in\{0\}\cup[N]$ are independent and identically distributed $\sigma$-sub-Gaussian noise.  
\end{assump}
Assumption~\ref{assump:obs_model} is reasonable as, in many applications, the performance value of a control system can be obtained through experiment or simulation with noise. However, the gradient or other higher-order information can be difficult to acquire.

Throughout the rest of this paper, we use the notation $X_t\defeq(x_1, x_2, \cdots, x_t)$ to define the sequence of sampled points up to step $t$. Therefore, the historical evaluations are 
\[
\mathcal{D}_{1:t}\defeq\left\{(x_\tau, y_{0,\tau}, \cdots, y_{N,\tau})\right\}_{\tau=1}^t.
\]
\begin{assump}
\label{assump:norm_kernel}
The kernel function is normalized, such that, $k_i(x, x)\leq1, \forall x\in\mathcal{X}, i\in\{0\}\cup[N]$, \rev{where $\mathcal{X}=\mathcal{Z}\times\Theta$}. 
\end{assump}
\begin{assump}
\label{assump:uniform_slater}
There exists $\xi>0$ independent of the context $z\in\mathcal{Z}$ and $\bar{\theta}(z)\in\Theta$, which can be dependent on $z$, such that $g(\bar{\theta}(z), z)\leq -\xi e$, where $e\in\mathbb{R}^N$ is the vector with all $1$s and the inequality is interpreted elementwise. 
\end{assump}
Asssumption~\ref{assump:uniform_slater} is the so-called Uniform Slater Condition. \finalRev{Assump.~\ref{assump:uniform_slater} is quite mild since we only require that a small $\xi$ exists rather than knowing a feasible solution as assumed in safe Bayesian optimization literature~\cite{sui2015safe}.} As an immediate implication of Assumption~\ref{assump:uniform_slater}, we have the following result. 
\begin{proposition}
\label{cor:Slater_dist}
There exists a probability distribution $\bar{\pi}(z)$ over $\Theta$, such that $\E_{\bar{\pi}(z)}\left(g(\theta, z)\right)\leq-\xi e$.~
\end{proposition}


\section{Performance metric}
We compare the controller parameters selected by our algorithm with the optimal tracking solution of problem~\eqref{eqn:prob_form}, given the context $z$ taking the value $z_t$ at step $t$. 

We are interested in two metrics, 
\begin{subequations}
\label{eqn:metric}
\begin{align}
\mathcal{R}_T&=\sum_{t=1}^T\left(f(\theta_t, z_t)-f(\theta^*(z_t), z_t)\right),\label{eqn:cumu_regret} \\
\mathcal{V}_T&=\left\|\left[\sum_{t=1}^Tg(\theta_t, z_t)\right]^+\right\|,\label{eqn:cumu_vio}
\end{align}
\end{subequations}
which are the cumulative regret compared to the optimal tracking solutions and the cumulative constraint violations. The regret here is essentially the \rev{\emph{contextual regret}~\cite{krause2011contextual}}, which is stronger than the regret considered in~\cite{zhou2022kernelized}.

The form of constraint violation in~\finalRev{\eqref{eqn:cumu_vio}} is the violation of cumulative constraint value. $\mathcal{V}_T/T$ gives the violation of the average constraint value, which is common in practice. For example, when $g$ represents some economic cost~(e.g., monetary expenses and energy consumption), it is usually of more interest to bound the average constraint value during a period rather than the whole constraint function sequence.

\section{Gaussian process regression}
As in the existing popular Bayesian optimization methods, we use Gaussian process surrogates to learn the unknown functions. As in~\cite{chowdhury2017kernelized}, we artificially introduce a Gaussian process $\mathcal{GP}(0, k_0(\cdot, \cdot))$ for the surrogate modelling of the unknown black-box function $f$. We also adopt an i.i.d Gaussian zero-mean noise model with noise variance $\lambda>0$, which can be chosen by the algorithm.

\finalRev{Recall that $x_t$ is defined as $(\theta_t, z_t)$. We} introduce the following functions of $(x, x^{\prime})$,
\begin{subequations}
\label{eq:mean_cov}
\begin{align}
\mu_{0,t}(x) &=k_{0}(x_{1:t}, x)^\top\left(K_{0,t}+\lambda I\right)^{-1} y_{0, 1: t}, \\\notag
k_{0,t}\left(x, x^{\prime}\right) &=k_{0}\left(x, x^{\prime}\right)\\
-&k_{0}(x_{1:t}, x)^\top\left(K_{0,t}+ \lambda I\right)^{-1} k_{0}\left(x_{1:t}, x^{\prime}\right), \\
\sigma_{0,t}^2(x) &=k_{0,t}(x, x)
\end{align}
\end{subequations}
with $k_{0}(x_{1:t}, x)=[k_0(x_1, x), k_0(x_2, x),\cdots, k_0(x_t, x)]^\top$, $K_{0,t}=(k_0(x,x^\prime))_{x,x^\prime\in X_t}$, $y_{0,1:t}=[y_{0,1}, y_{0, 2},\cdots,y_{0,t}]^\top$. Similarly, we can get $\mu_{i,t}(\cdot), k_{i,t}(\cdot, \cdot), \sigma_{i,t}(\cdot)$, $\forall i\in[N]$ for the constraints.

To characterize the complexity of the Gaussian processes and the RKHSs corresponding to the kernel functions, we further introduce the maximum information gain for the objective $f$ as in~\cite{srinivas2012information},
\begin{equation}
\label{eq:max_inf_gain}
\gamma_{0,t}:=\max_{A \subset \Theta\times\mathcal{Z};|A|=t} \frac{1}{2} \log \left|I+\lambda^{-1}K_{0,A}\right|,
\end{equation}
where $K_{0,A}=(k_0(x, x^\prime))_{x,x^\prime\in A}$. 
Similarly, we introduce $\gamma_{i,t},\forall i\in[N]$ for the constraints. 
\begin{myRemark}
Note that the Gaussian process model above is \emph{only} used to derive posterior mean functions, covariance functions, and maximum information gain for algorithm design and theoretical analysis. It does not change our set-up that $f$ is a deterministic function and that the observation noise only needs to be sub-Gaussian. 
\end{myRemark}

Based on the aforementioned preliminaries of Gaussian process regression, we then derive the lower confidence, and upper confidence bound functions.
\begin{lemma}
\label{lem:hpb_int}
Let Assumptions~\ref{assump:support_set} and \ref{assump:bounded_norm} hold. 
With probability at least $1-\delta,\forall\delta\in(0, 1)$, the following holds for all $x\in\mathcal{X}$ and $t\geq1$,
\begin{align}
g_i(x)&\in[\lowg{i,t}(x),\ug{i,t}(x)],\;\forall i\in[N] \\
\mathrm{and }\;\;f(x)&\in[\lf{t}(x),\uf{t}(x)]\;,
\end{align}
where for all $i\in[N]$,
\begin{subequations}
\label{eqn:lu_defs}
\begin{align}
\lf{t}(x)&\defeq\max\{\mu_{0,t-1}(x)-\beta^{1/2}_{0,t}\sigma_{0,t-1}(x), -C_0\}\enspace,\\
\uf{t}(x)&\defeq\min\{\mu_{0,t-1}(x)+\beta^{1/2}_{0,t}\sigma_{0,t-1}(x), C_0\}\enspace,\\
\lowg{i,t}(x)&\defeq\max\{\mu_{i,t-1}(x)-\beta^{1/2}_{i,t}\sigma_{i,t-1}(x), -C_i\}\enspace,\\
\ug{i,t}(x)&\defeq\min\{\mu_{i,t-1}(x)+\beta^{1/2}_{i,t}\sigma_{i,t-1}(x), C_i\}\enspace,\\
\beta^{1/2}_{i,t}&\defeq C_i+\sigma \sqrt{2\left(\gamma_{i,t-1}+1+\ln ((N+1) / \delta)\right)}\enspace.
\end{align}
\end{subequations}
\end{lemma}
\begin{proof}
\finalRev{By Corollary 2.6,~\cite{xu2023constrained}, with probability at least $1-\delta,\forall\delta\in(0, 1)$, for all $x\in\mathcal{X}$ and $t\geq1$,
$$
\mu_{i,t-1}(x)-\beta^{1/2}_{i,t}\sigma_{i,t-1}(x)\leq g_i(x)\leq \mu_{i,t-1}(x)+\beta^{1/2}_{i,t}\sigma_{i,t-1}(x).
$$
Furthermore, $|g_i(x)|=|\langle g_i, k_i(x, \cdot)\rangle|\leq \|g_i\|\|k_i(x, \cdot)\|\leq C_i, \forall i\in[N]$. Therefore, $g_i(x)\in[\lowg{i,t}(x),\ug{i,t}(x)]$. Similarly, $f(x)\in[\lf{t}(x),\uf{t}(x)]$.
}
\end{proof}

Without further notice, all the following results are conditioned on the high probability event in Lem.~\ref{lem:hpb_int} happening.

\section{Algorithm and Theoretical Guarantees}

Our primal-dual algorithm is shown in Alg.~\ref{alg:pdcbo}, where \finalRev{$\eta$ is a small scaling factor to be chosen, $\lambda_t$ can be interpreted as the dual variable up to the scaling of $\eta$}, $0<\epsilon\leq\frac{\xi}{2}$ is a slackness parameter\finalRev{,} and $[\cdot]^+$ is interpreted element-wise. \finalRev{Intuitively, the larger $\eta$ is, the more emphasis is given to the constraints.} 
\begin{algorithm}[htbp!]
\caption{\textbf{P}rimal-\textbf{D}ual \textbf{C}ontextual \textbf{B}ayesian \textbf{O}ptimization~(PDCBO).}
\label{alg:pdcbo}
\begin{algorithmic}[1]
\normalsize
\For{$t\in[T]$}
\State Observe contextual variables $z_t$. 
\State \textbf{Primal update:} \begin{equation}\label{eqn:primal_udt}\theta_{t}=\arg\min_{\theta\in\Theta}\{\lf{t}(\theta, z_{t})+\eta\lambda_t^T\lowg{t}(\theta, z_t)\}.\end{equation}
\vspace{-3ex}
 \State \textbf{Dual update:} \begin{equation}\label{eqn:dual_udt}\lambda_{t+1}=[\lambda_t+ \lowg{t}(\theta_{t}, z_t)+\epsilon e]^+.\end{equation}
 \vspace{-3ex}
 \State \rev{Evaluate} $f$ and $g_i,i\in[N]$ at $x_t=(\theta_t, z_t)$ \rev{with noise}. \label{alg_line:eval}
 \State \rev{Update $(\mu_{i,t}, \sigma_{i,t}), i\in\{0\}\cup[N]$ with the new data.} 
\EndFor
\end{algorithmic}
\end{algorithm}
\vspace{-0.3cm}

\rev{The Alg.~\ref{alg:pdcbo} is based on the well-known primal-dual optimization method. The Lagrangian of the original problem is $\mathcal{L}(\theta, z, \phi)=f(\theta, z)+\phi^Tg(\theta, z)$ and the dual function is $\mathcal{D}(z, \phi)=\min_{\theta}\mathcal{L}(\theta,z,\phi)$. In the primal update step, we optimize the Lagrangian. In the dual update step, we apply the dual ascent method. In primal and dual updates, we replace the unknown black-box functions with their lower confidence bounds, following the principle of \emph{optimism in the face of uncertainty}. } 

\rev{The primal update~\eqref{eqn:primal_udt} requires solving a potentially non-convex auxiliary optimization problem. When the dimension of $\theta$ is small~(e.g., $<6$) we can use a pure grid search method to solve it. For larger dimensions, one can, for example, apply the gradient descent method from multiple different initial points.} \finalRev{{Line~\ref{alg_line:eval} can be done by querying a simulator or doing an experiment in practice.}}
\begin{remark}
\rev{In Alg.~\ref{alg:pdcbo}}, $\lambda_t$ is not \rev{exactly} the dual variable, but the dual variable scaled by $\frac{1}{\eta}$. Indeed, $\lambda_t$ can be interpreted as virtual queue length~\cite{zhou2022kernelized}. The intuition of $\epsilon$ is to introduce some level of constant pessimistic drift to control the cumulative violation.
\end{remark}

\rev{We now give the theoretical guarantees in Thm.~\ref{thm:main_thm}.}
\begin{theorem}
\label{thm:main_thm}
Let the  Assumptions~\ref{assump:support_set},~\ref{assump:bounded_norm},~\ref{assump:obs_model},~\ref{assump:norm_kernel} and~\ref{assump:uniform_slater} hold. \rev{We further assume the maximum information gain term satisfies $\lim_{T\to\infty}{\frac{{{\|\gamma_T^g\|}}}{\sqrt{T}}}=0$, where $\gamma_T^g\defeq(\gamma_{1,T}, \cdots, \gamma_{N,T})$.} We set $\eta=\frac{1}{\sqrt{T}}$, $\lambda_1=(\frac{4C_0}{\eta\xi}+\frac{4\|C\|^2}{\xi})e$ and 
{\small
\[
\epsilon=\frac{\sqrt{{N}\left(\frac{4C_0}{\eta\xi}+\frac{4\|C\|^2}{\xi}\right)^2+\frac{4C_0}{\eta}+4\|C\|^2}+8\|\beta_T^g\|\sqrt{T\|\gamma^g_T\|}}{T},
\]
} where $\beta_T^g\defeq(\beta_{1,T},\cdots,\beta_{N,T})$ \rev{and $C\defeq(C_1,\cdots,C_N)$}. Let $T$ be large enough such that $\epsilon=\mathcal{O}(\|\gamma_T^g\|/\sqrt{T})\leq\xi/2$. We have 
\[
\mathcal{R}_T=\mathcal{O}\left(\sum_{i=0}^N\gamma_{i,T}\sqrt{T}\right)\;\;
\mathrm{and}\;\;\mathcal{V}_T\finalRev{=}0.
\]
\end{theorem}
\emph{\textbf{Proof Sketch}} Our proof involves three key ingredients. First, the cumulative cost paid to learn the black-box functions can be bounded by some term involving the maximum information gain~\eqref{eq:max_inf_gain}. Second, the regret and \rev{the} violations in the primal values can be bounded by the drift of dual variables. Third, by properly selecting the constant dual drift $\epsilon$~\footnote{\finalRev{In practice, we can set $\epsilon$ to be a small value as compared to the range of the constraint function or even just $0$.}}, we can balance the regret and violation such that \emph{time-average} constraints are strictly satisfied while the cumulative \emph{contextual regret} is bounded. \rev{The detailed proofs can be found in the Appendix \shortversion{of our technical report~\cite{xu2023primal}}.}  

\begin{remark}
Thm.~\ref{thm:main_thm} highlights that for kernels with \rev{the maximum} information \rev{gain} terms that grow slower than $\sqrt{T}$, our algorithm achieves average \rev{contextual} regret converging to zero while incurring no constraint violation on average. \finalRev{The slow growth assumption holds for most popular kernels including Squared Exponential and M\'atern~(under the assumption shown in Tab.~\ref{tab:kern_spec_bounds}) kernels~\cite{srinivas2012information}.} 
\end{remark}
\rev{By applying the bounds on maximum information gains from~\cite{srinivas2012information,vakili2021information} to Thm.~\ref{thm:main_thm}, we derive the kernel-specific bounds as in Tab.~\ref{tab:kern_spec_bounds}, \rev{where $d=n_\theta+n_z$, $\nu$ is the smoothness parameter of the M\'atern kernel, and $\tilde{\mathcal{O}}(\cdot)$ hides some polylogarithmic term}. 
\begin{table}[htbp]
    \caption{Kernel-specific regret bounds.}
    \label{tab:kern_spec_bounds}
    \scalebox{0.9}{\begin{tabular}{|c|c|c|c|}
      \hline 
       Kernel &  Linear & Squared Exponential & M\'atern~($\frac{\nu}{d}>\frac{1}{2}$) \\
       \hline 
       $\mathcal{R}_T$ & $\tilde{\mathcal{O}}(d\sqrt{T})$  & $\tilde{\mathcal{O}}(\sqrt{T})$ &$\tilde{\mathcal{O}}(T^{\frac{2\nu+3d }{4 \nu+2d}})$ \\
       \hline 
    \end{tabular}
    }
\end{table}
}
\begin{remark}
\rev{In Thm.~\ref{thm:main_thm}, the running steps $T$ is known beforehand. To extend to the case where $T$ is unknown, we can apply the doubling trick~\cite{besson2018doubling}. The idea is to start the algorithm with a small $T$ and restart it with $T$ doubled every time the running steps are exhausted.} 
\end{remark}

\section{Experiments}
We consider two sets of experiments. In the first set, we use the objective and constraint functions sampled from Gaussian processes. In the second set, we consider the classical Williams-Otto benchmark problem~\cite{del2021real}. \rev{We compare our method to the state-of-the-art Bayesian optimization methods with constraints consideration, including \textsf{SafeOPT}~\finalRev{\cite{sui2015safe,berkenkamp2021bayesian}} and \textsf{CEI}~(\textbf{C}onstrained \textbf{E}xpected \textbf{I}mprovement) method~\cite{gelbart2014bayesian,gardner2014bayesian}. The experiments are implemented in \textsf{python}, based on the package \textsf{GPy}~\cite{gpy2014}.} \finalRev{The code for the experiments is available through \href{https://github.com/PREDICT-EPFL/PDCBO}{https://github.com/PREDICT-EPFL/PDCBO}.} 

\rev{\textbf{Choice of Parameters.} The performance of PDCBO is mainly impacted by $\beta^{1/2}_{i,t}$. In practice, $\beta^{1/2}_{i,t}$ can usually be set as a constant. \finalRev{Indeed, when the kernel choices and the kernel hyperparameters fit the black-box functions well, setting $\beta^{1/2}_{i,t}=3$ typically works well.} In our experiments, manually setting $\beta^{1/2}_{i,t}=1.0$ works well. We also set $\lambda=0.05^2$ as the noise variance for the Gaussian process modeling. We use the common squared exponential kernel. For the second example, we sample a few points randomly and maximize the likelihood function to get the hyperparameters of the kernel.}  

\rev{\textbf{Computational Time.} In our experiments, all the problems have low-dimensional inputs~($n_\theta\leq3$). So we use pure grid search to solve the auxiliary problem for primal update, which is relatively cheap due to the explicit expressions of the lower confidence bound functions as compared to the simulation of the Williams-Otto benchmark problem.
}  

\finalRev{\textbf{Performance Metrics.} To measure the quality of sample sequences, we use constrained contextual regret shown in Eq.~\ref{eqn:cumu_regret}. To measure the violations, we introduce the cumulative constraint value \finalRev{$\sum_{\tau=1}^{t}g_i(\theta_\tau, z_\tau), i\in[N]$,} and the time-average constraint value $\frac{1}{t}\sum_{\tau=1}^{t}g_i(\theta_\tau, z_\tau), i\in[N]$.  
}

\subsection{Sampled Instances from Gaussian Process}
We consider the problem~\eqref{eqn:prob_form} with only one constraint, where $\theta\in\Theta=[-10, 10]$, $z\in\mathcal{Z}=[-10, 10]$, and both $f$ and $g$ are unknown black-box functions sampled from a Gaussian process. We use the squared exponential kernel,
\begin{align}
    k((\theta_1, z_1), (\theta_2, z_2))=&\\\notag
    \sigma_\textrm{SE}^2\exp&\left\{-\left(\frac{\theta_1-\theta_2}{l_\theta}\right)^2-\left(\frac{z_1-z_2}{l_z}\right)^2\right\},
\end{align}
where $\sigma_\textrm{SE}^2=2.0$, and $l_\theta=l_z=1.0$. 
\begin{figure}[htbp!]
    \centering
    \input{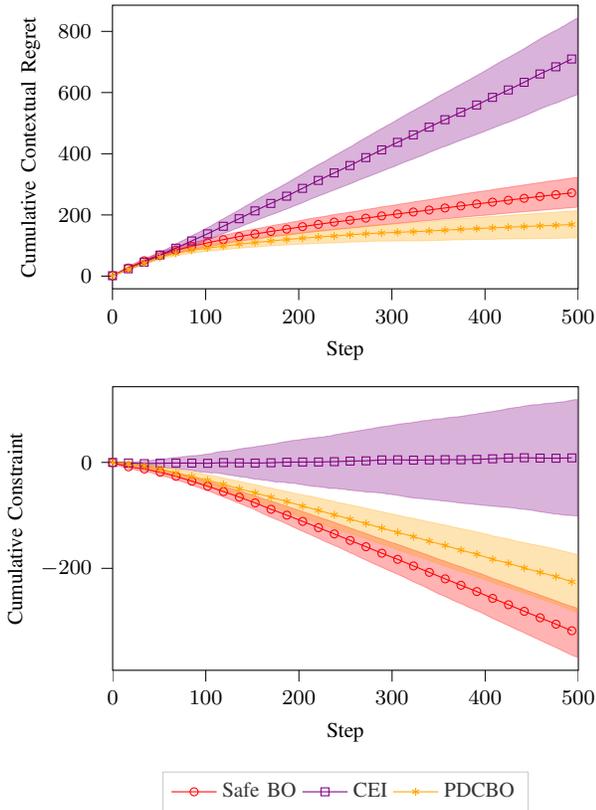}
    \caption{The cumulative context regret and cumulative constraint value of different algorithms. The statistics are obtained over 50 different trajectories. The shaded area represents $\pm 0.5$ standard deviation.}
    \label{fig:cumu_dyr_and_constrs}
\end{figure}
We run different algorithms for $500$ steps over $50$ different sampled instances. The context sequence is randomly generated from a uniform distribution over the context set $\mathcal{Z}$. Fig.~\ref{fig:cumu_dyr_and_constrs} shows the cumulative contextual regret and cumulative constraint value of different algorithms over $50$ different trajectories. It can be observed that our PDCBO 
achieves sublinear cumulative contextual regret while keeping the cumulative constraint value below zero with high confidence. In contrast, the CEI method suffers from almost linear cumulative contextual regret and fails to keep average constraint feasibility with high probability. Although Safe BO can keep average constraint feasibility, its cumulative contextual regret is about $62\%$ higher than our PDCBO due to its being overly cautious to keep feasibility in every single step.

\subsection{Williams-Otto Classical Problem}
In this section, we consider the problem of the classical Williams-Otto benchmark problem~\cite{del2021real}. In this problem, we feed a continuous stirred-tank reactor~(CSTR) with two components \chemfig{A} and \chemfig{B}. The reactor then operates at the steady state under the reaction temperature $T_\mathrm{r}$. During the chemical reaction, a byproduct \chemfig{G} is produced. We use $X_\mathrm{A}$ and $X_\mathrm{G}$ to denote the steady-state residual mass fractions of \chemfig{A} and \chemfig{G} at the reactor outlet. Both $X_\mathrm{A}$ and $X_\mathrm{G}$ need to be managed. That is, we need to control the steady-state residual mass fractions of \chemfig{A} and \chemfig{G} to be lower than some thresholds. See \cite[Sec.~4.1]{del2021real} for more details. Meanwhile, a key factor that impacts the cumulative profit of the chemical reaction process is the time-varying nature of the product price and the costs of the raw materials~\cite{yadbantung2022periodically}. To incorporate these time-varying factors, we model them as contextual variables.  

We want to adaptively tune the feed rate $F_\mathrm{B}$ of the component \chemfig{B} and the reaction temperature $T_\mathrm{r}$ to maximize the cumulative economic profit from the reaction while managing $X_\mathrm{A}$ and $X_\mathrm{G}$. Our problem can be formulated as, 
\begin{equation}
\begin{aligned}
\min_{F_\mathrm{B}, T_{\mathrm{r}}}\quad& J(F_\mathrm{B}, T_\mathrm{r}, P)\\
\text{subject to}\quad &\textrm{CSTR model~\cite{mendoza2016assessing}}\\
& g_1(F_\mathrm{B}, T_\mathrm{r})\defeq X_\mathrm{A}(F_\mathrm{B}, T_\mathrm{r})-0.12 \leq 0\\
&  g_2(F_\mathrm{B}, T_\mathrm{r})\defeq X_\mathrm{G}(F_\mathrm{B}, T_\mathrm{r})-0.08 \leq 0,\\
& F_\mathrm{B}\in[4, 7], T_\mathrm{r}\in[70, 100]\\ 
\end{aligned}
\end{equation}
where $J(F_\mathrm{B}, T_\mathrm{r}, P)$ is the minimization objective that is opposite to the net economic profit, $P\in\mathbb{R}^4$ is the price vector of the product and the raw materials, $g_1(F_\mathrm{B}, T_\mathrm{r})$ and $g_2(F_\mathrm{B}, T_\mathrm{r})$ are threshold constraints on the residual mass fractions. To generate the context $P^t$ at step $t$, we randomly sample $P^t_i$ uniformly from the interval \finalRev{$[(1-\alpha)\bar{P}_i, (1+\alpha)\bar{P}_i]$, where $\bar{P}_i$ is a predefined normal price and $\alpha=0.2$}. 
\begin{figure}[htbp!]
    \centering    
    \input{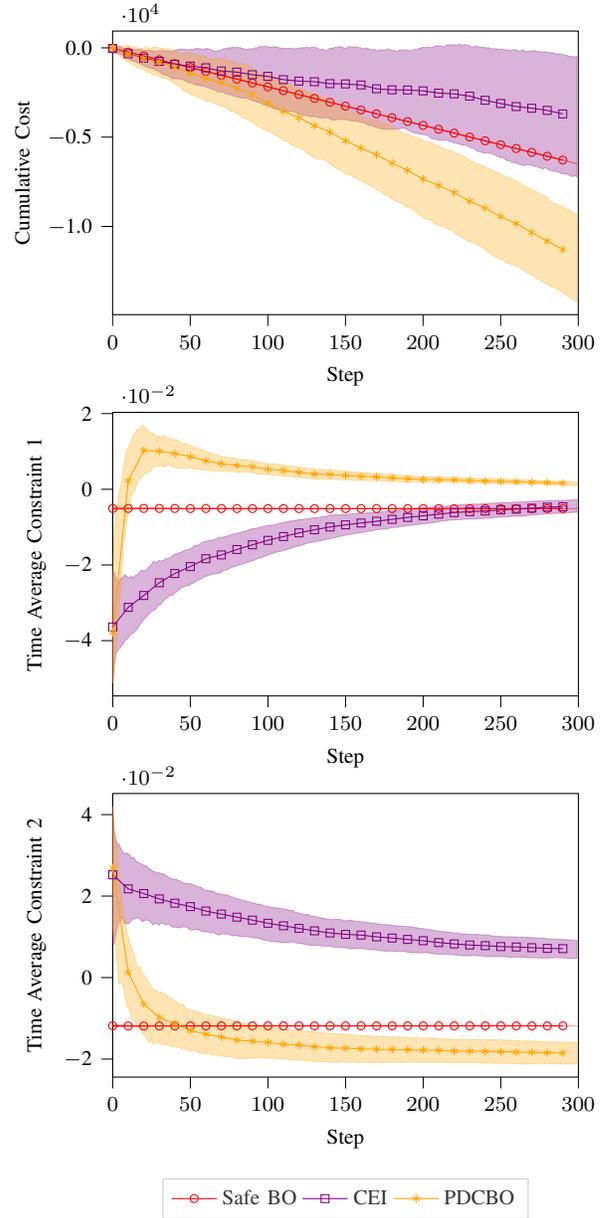}
    \caption{Cumulative cost and average constraints of the chemical reaction over $50$ different context trajectories. The shaded area represents $\pm1.0$ standard deviation.}
    \label{fig:cstr_res}
\end{figure}

Fig.~\ref{fig:cstr_res} shows the cumulative cost and the running average of the constraints with 50 different context trajectories. We observe that our PDCBO achieves the lowest cumulative cost (the highest cumulative profit, equivalently) while satisfying the constraints on average. In contrast, Safe BO is overly cautious and can not achieve all the potential economic profit. Finally, the generic CEI method suffers from both low profit and severe constraint violation for the second constraint.  

\section{Conclusion}
This paper studied the problem of optimizing the closed-loop performance of constrained control systems, where both the objective and the constraints are unknown black-box functions affected by exogenous time-varying contextual disturbances. A primal-dual contextual Bayesian optimization algorithm was proposed, and it was shown to achieve sublinear cumulative regret with respect to the dynamic optimal solution under certain regularity conditions. Furthermore, the algorithm achieves zero time-average constraint violation, ensuring that the average value of the constraint function satisfies the desired constraint.
Our method is particularly useful when the constraint functions correspond to some economic costs that accumulate with time. The method is applied to both sampled instances from Gaussian processes and to a continuous stirred tank reactor \rev{feedrate and reaction temperature} tuning problem; simulation results on both problems show that the method simultaneously provides close-to-optimal performance and maintains constraint feasibility on average. This contrasts current state-of-the-art methods, which either suffer from large cumulative regret or severe constraint violations for the case studies presented.




\bibliographystyle{IEEEtran}
\bibliography{refs}

\begin{thebibliography}{10}
\providecommand{\url}[1]{#1}
\csname url@samestyle\endcsname
\providecommand{\newblock}{\relax}
\providecommand{\bibinfo}[2]{#2}
\providecommand{\BIBentrySTDinterwordspacing}{\spaceskip=0pt\relax}
\providecommand{\BIBentryALTinterwordstretchfactor}{4}
\providecommand{\BIBentryALTinterwordspacing}{\spaceskip=\fontdimen2\font plus
\BIBentryALTinterwordstretchfactor\fontdimen3\font minus
  \fontdimen4\font\relax}
\providecommand{\BIBforeignlanguage}[2]{{%
\expandafter\ifx\csname l@#1\endcsname\relax
\typeout{** WARNING: IEEEtran.bst: No hyphenation pattern has been}%
\typeout{** loaded for the language `#1'. Using the pattern for}%
\typeout{** the default language instead.}%
\else
\language=\csname l@#1\endcsname
\fi
#2}}
\providecommand{\BIBdecl}{\relax}
\BIBdecl

\bibitem{xu2022vabo}
W.~Xu, C.~N. Jones, B.~Svetozarevic, C.~R. Laughman, and A.~Chakrabarty,
  ``{VABO}: Violation-{A}ware {B}ayesian {O}ptimization for closed-loop control
  performance optimization with unmodeled constraints,'' in \emph{2022 American
  Control Conference (ACC)}.\hskip 1em plus 0.5em minus 0.4em\relax IEEE, 2022,
  pp. 5288--5293.

\bibitem{xu2022lower}
W.~Xu, Y.~Jiang, E.~T. Maddalena, and C.~N. Jones, ``Lower bounds on the
  worst-case complexity of efficient global optimization,'' \emph{arXiv
  preprint arXiv:2209.09655}, 2022.

\bibitem{williams2006gaussian}
C.~K. Williams and C.~E. Rasmussen, \emph{Gaussian processes for machine
  learning}.\hskip 1em plus 0.5em minus 0.4em\relax MIT press Cambridge, MA,
  2006, vol.~2, no.~3.

\bibitem{gelbart2014bayesian}
M.~A. Gelbart, J.~Snoek, and R.~P. Adams, ``Bayesian optimization with unknown
  constraints,'' in \emph{Proc. of the 30th Conf. on Uncertainty in Artif.
  Intell.}, ser. UAI'14.\hskip 1em plus 0.5em minus 0.4em\relax Arlington,
  Virginia, USA: AUAI Press, 2014, p. 250–259.

\bibitem{gardner2014bayesian}
J.~R. Gardner, M.~J. Kusner, Z.~E. Xu, K.~Q. Weinberger, and J.~P. Cunningham,
  ``Bayesian optimization with inequality constraints.'' in \emph{Proc. of the
  Int. Conf. on Mach. Learn.}, vol. 2014, 2014, pp. 937--945.

\bibitem{sui2015safe}
Y.~Sui, A.~Gotovos, J.~Burdick, and A.~Krause, ``Safe exploration for
  optimization with {G}aussian processes,'' in \emph{Proc. of the Int. Conf. on
  Mach. Learn.}, 2015, pp. 997--1005.

\bibitem{sui2018stagewise}
Y.~Sui, J.~Burdick, Y.~Yue \emph{et~al.}, ``Stage-wise safe {B}ayesian
  optimization with {G}aussian processes,'' in \emph{Proc. of the Int. Conf. on
  Mach. Learn.}, 2018, pp. 4781--4789.

\bibitem{bergmann2020safe}
D.~Bergmann and K.~Graichen, ``Safe {B}ayesian optimization under unknown
  constraints,'' in \emph{2020 59th IEEE Conference on Decision and Control
  (CDC)}.\hskip 1em plus 0.5em minus 0.4em\relax IEEE, 2020, pp. 3592--3597.

\bibitem{baumann2021gosafe}
D.~Baumann, A.~Marco, M.~Turchetta, and S.~Trimpe, ``Gosafe: Globally optimal
  safe robot learning,'' in \emph{2021 IEEE International Conference on
  Robotics and Automation (ICRA)}.\hskip 1em plus 0.5em minus 0.4em\relax IEEE,
  2021, pp. 4452--4458.

\bibitem{rupenyan2021performance}
A.~Rupenyan, M.~Khosravi, and J.~Lygeros, ``Performance-based trajectory
  optimization for path following control using {B}ayesian optimization,'' in
  \emph{2021 60th IEEE Conference on Decision and Control (CDC)}.\hskip 1em
  plus 0.5em minus 0.4em\relax IEEE, 2021, pp. 2116--2121.

\bibitem{fiducioso2019safe}
M.~Fiducioso, S.~Curi, B.~Schumacher, M.~Gwerder, and A.~Krause, ``Safe
  contextual {B}ayesian optimization for sustainable room temperature {P}{I}{D}
  control tuning,'' in \emph{Proc. of the Twenty-Eighth Int. Joint Conf. on
  Artif. Intell.}\hskip 1em plus 0.5em minus 0.4em\relax IJCAI, 2019, pp.
  5850--5856.

\bibitem{brunzema2022controller}
P.~Brunzema, A.~Von~Rohr, and S.~Trimpe, ``On controller tuning with
  time-varying {B}ayesian optimization,'' in \emph{2022 IEEE 61st Conference on
  Decision and Control (CDC)}.\hskip 1em plus 0.5em minus 0.4em\relax IEEE,
  2022, pp. 4046--4052.

\bibitem{bogunovic2016time}
I.~Bogunovic, J.~Scarlett, and V.~Cevher, ``Time-varying {G}aussian process
  bandit optimization,'' in \emph{Artificial Intelligence and
  Statistics}.\hskip 1em plus 0.5em minus 0.4em\relax PMLR, 2016, pp. 314--323.

\bibitem{deng2022weighted}
Y.~Deng, X.~Zhou, B.~Kim, A.~Tewari, A.~Gupta, and N.~Shroff, ``Weighted
  {G}aussian process bandits for non-stationary environments,'' in
  \emph{International Conference on Artificial Intelligence and
  Statistics}.\hskip 1em plus 0.5em minus 0.4em\relax PMLR, 2022, pp.
  6909--6932.

\bibitem{xu2023violation}
W.~Xu, C.~N. Jones, B.~Svetozarevic, C.~R. Laughman, and A.~Chakrabarty,
  ``Violation-aware contextual {B}ayesian optimization for controller
  performance optimization with unmodeled constraints,'' \emph{arXiv preprint
  arXiv:2301.12099}, 2023.

\bibitem{fleming2017time}
J.~Fleming and M.~Cannon, ``Time-average constraints in stochastic {M}odel
  {P}redictive {C}ontrol,'' in \emph{2017 American Control Conference
  (ACC)}.\hskip 1em plus 0.5em minus 0.4em\relax IEEE, 2017, pp. 5648--5653.

\bibitem{korda2014stochastic}
M.~Korda, R.~Gondhalekar, F.~Oldewurtel, and C.~N. Jones, ``Stochastic {MPC}
  framework for controlling the average constraint violation,'' \emph{IEEE
  Transactions on Automatic Control}, vol.~59, no.~7, pp. 1706--1721, 2014.

\bibitem{cannon2009probabilistic}
M.~Cannon, B.~Kouvaritakis, and X.~Wu, ``Probabilistic constrained {MPC} for
  multiplicative and additive stochastic uncertainty,'' \emph{IEEE Transactions
  on Automatic Control}, vol.~54, no.~7, pp. 1626--1632, 2009.

\bibitem{kim2012free}
J.~Kim, M.~Ruggiero, and D.~Atienza, ``Free cooling-aware dynamic power
  management for green datacenters,'' in \emph{2012 International Conference on
  High Performance Computing \& Simulation (HPCS)}.\hskip 1em plus 0.5em minus
  0.4em\relax IEEE, 2012, pp. 140--146.

\bibitem{zhou2022kernelized}
X.~Zhou and B.~Ji, ``On kernelized multi-armed bandits with constraints,'' in
  \emph{Advances in Neural Information Processing Systems}, 2022.

\bibitem{krause2011contextual}
A.~Krause and C.~Ong, ``Contextual {G}aussian process bandit optimization,''
  \emph{Advances in Neural Information Processing Systems}, vol.~24, 2011.

\bibitem{scholkopf2001generalized}
B.~Sch{\"o}lkopf, R.~Herbrich, and A.~J. Smola, ``A generalized representer
  theorem,'' in \emph{International conference on computational learning
  theory}.\hskip 1em plus 0.5em minus 0.4em\relax Springer, 2001, pp. 416--426.

\bibitem{srinivas2012information}
N.~Srinivas, A.~Krause, S.~M. Kakade, and M.~W. Seeger, ``Information-theoretic
  regret bounds for {G}aussian process optimization in the bandit setting,''
  \emph{IEEE Transactions on Information Theory}, vol.~58, no.~5, pp.
  3250--3265, 2012.

\bibitem{chowdhury2017kernelized}
S.~R. Chowdhury and A.~Gopalan, ``On kernelized multi-armed bandits,'' in
  \emph{Int. Conf. on Mach. Learn.}\hskip 1em plus 0.5em minus 0.4em\relax
  PMLR, 2017, pp. 844--853.

\bibitem{xu2023constrained}
W.~Xu, Y.~Jiang, B.~Svetozarevic, and C.~Jones, ``Constrained efficient global
  optimization of expensive black-box functions,'' in \emph{International
  Conference on Machine Learning}.\hskip 1em plus 0.5em minus 0.4em\relax PMLR,
  2023, pp. 38\,485--38\,498.

\bibitem{vakili2021information}
S.~Vakili, K.~Khezeli, and V.~Picheny, ``On information gain and regret bounds
  in {G}aussian process bandits,'' in \emph{International Conference on
  Artificial Intelligence and Statistics}.\hskip 1em plus 0.5em minus
  0.4em\relax PMLR, 2021, pp. 82--90.

\bibitem{besson2018doubling}
L.~Besson and E.~Kaufmann, ``What doubling tricks can and can't do for
  multi-armed bandits,'' \emph{arXiv preprint arXiv:1803.06971}, 2018.

\bibitem{del2021real}
E.~A. del Rio~Chanona, P.~Petsagkourakis, E.~Bradford, J.~A. Graciano, and
  B.~Chachuat, ``Real-time optimization meets {B}ayesian optimization and
  derivative-free optimization: {A} tale of modifier adaptation,''
  \emph{Comput. \& Chem. Eng.}, vol. 147, p. 107249, 2021.

\bibitem{berkenkamp2021bayesian}
F.~Berkenkamp, A.~Krause, and A.~P. Schoellig, ``Bayesian optimization with
  safety constraints: safe and automatic parameter tuning in robotics,''
  \emph{Machine Learning}, pp. 1--35, 2021.

\bibitem{gpy2014}
{GPy}, ``{GPy}: A {G}aussian process framework in python,''
  \url{http://github.com/SheffieldML/GPy}, since 2012.

\bibitem{yadbantung2022periodically}
R.~Yadbantung and P.~Bumroongsri, ``Periodically time-varying economic model
  predictive control with applications to nonlinear continuous stirred tank
  reactors,'' \emph{Computers \& Chemical Engineering}, vol. 157, p. 107602,
  2022.

\bibitem{mendoza2016assessing}
D.~F. Mendoza, J.~E.~A. Graciano, F.~dos Santos~Liporace, and G.~A.~C. Le~Roux,
  ``Assessing the reliability of different real-time optimization
  methodologies,'' \emph{The Can. J. of Chem. Eng.}, vol.~94, no.~3, pp.
  485--497, 2016.

\bibitem{chowdhury2017kernelized_arxiv}
S.~R. Chowdhury and A.~Gopalan, ``On kernelized multi-armed bandits,''
  \emph{arXiv preprint arXiv:1704.00445}, 2017.

\end{thebibliography}

\longversion{
\appendix 
\rev{We give the detailed proof of the Thm.~\ref{thm:main_thm}. Note that all the results are under the same assumptions as the Thm.~\ref{thm:main_thm} without further notice.} The learning cost comes from the uncertainties of the unknown functions, measured by the posterior standard deviations, and can be bounded as in Lem.~\ref{lem:bound_cumu_sd}. 
\begin{lemma}[Lemma 4,~\cite{chowdhury2017kernelized_arxiv}]
\label{lem:bound_cumu_sd}
Given a sequence of points $x_1, x_2, \cdots, x_T$ from $\Theta\times\mathcal{Z}$, we have,
\begin{equation}
\sum_{t=1}^T \sigma_{i, t-1}\left(x_t\right) \leq \sqrt{4(T+2) \gamma_{i,T}}.
\end{equation}
\end{lemma}
To connect \rev{the drift of the dual variables} and the primal values, we introduce a function in the dual space, 
\begin{equation}
V(\lambda_t) = \frac{1}{2}\|\lambda_t\|^2.
\end{equation}
We consider,
\begin{subequations}
\label{eqn:dual_dec_ineq}
\begin{align}
\Delta_t \defeq& V(\lambda_{t+1})-V(\lambda_t)\\
=& \frac{1}{2}\left(\|[\lambda_t+\lowg{t}(\theta_t, z_t)+\epsilon e]^+\|^2-\|\lambda_t\|^2\right)\\
\leq&\frac{1}{2}\left(\|\lambda_t+\lowg{t}(\theta_t, z_t)+\epsilon e\|^2-\|\lambda_t\|^2\right)\\
=& \lambda_t^T(\lowg{t}(\theta_t, z_t)+\epsilon e) + \frac{1}{2}\|\lowg{t}(\theta_t, z_t)+\epsilon e\|^2,
\end{align}
\end{subequations}
\rev{where the inequality follows by discussion on the sign of $\lambda_t+\lowg{t}(\theta_t, z_t)+\epsilon e$.} To characterize the tradeoff between feasibility and optimality, we introduce the perturbed problem,
\begin{subequations}
\label{eqn:rel_epsilon_prob_form}
\begin{align}
\min_{\pi\in\Pi(\Theta)}  \quad& {\E}_\pi(f(\theta, z)),\label{eqn:rel_epsilon_obj} \\
\text{subject to:} \quad & {\E}_\pi (g(\theta, z))+\epsilon e\leq 0,\quad \forall i\in[N],\label{eqn:rel_epsilon_constraint}
\end{align}
\end{subequations}
\rev{where the feasible set is relaxed to the set of all distributions over the set $\Theta$. Such a relaxation results in a linear programming problem in distribution, which is easier for sensitivity analysis.}
We use $\pi^*_\epsilon(z)$ to denote the optimal solution to the above problem. We then have the following lemma.
\begin{lemma}
\label{lem:bound_pert_opt} 
\begin{equation}
\sum_{t=1}^T\E_{\pi^*_\epsilon(z_t)}(f(\theta, z_t))-\sum_{t=1}^T\E_{\pi^*(z_t)}(f(\theta, z_t))\leq\frac{2C_0T\epsilon}{\xi},\end{equation} where $\pi^*(z_t)$ is the optimal distribution for Problem~\eqref{eqn:rel_epsilon_prob_form} with $\epsilon=0$. 
\end{lemma}
\begin{proof}
Let $\pi_\epsilon(z_t)=(1-\frac{\epsilon}{\xi})\pi^*(z_t)+\frac{\epsilon}{\xi}\bar{\pi}(z_t)$, where $\E_{\bar{\pi}(z_t)}(g(\theta, z_t))\leq-\xi e$~(Recall the Proposition~\ref{cor:Slater_dist}). Then 
\begin{align}
\E_{\pi_\epsilon(z_t)}(g(\theta,z_t))&=(1-\frac{\epsilon}{\xi})\E_{\pi^*(z_t)}(g(\theta,z_t))+\frac{\epsilon}{\xi}\E_{\bar{\pi}(z_t)}(g(\theta, z_t))\nonumber\\ 
&\leq-\epsilon e. \nonumber
\end{align}
Hence, $\pi_\epsilon(z_t)$ is a feasible solution to the slightly perturbed problem. So, we have,
\begin{align*}
&\sum_{t=1}^T\E_{\pi^*_\epsilon(z)}(f(\theta, z_t))-\sum_{t=1}^T\E_{\pi^*(z)}(f(\theta, z_t))\\
\leq&\sum_{t=1}^T\E_{\pi_\epsilon(z)}(f(\theta, z_t))-\sum_{t=1}^T\E_{\pi^*(z)}(f(\theta, z_t))\\
=&\frac{\epsilon}{\xi}\sum_{t=1}^T\left(\E_{\bar{\pi}(z_t)}(f(\theta, z_t))-\E_{\pi^*(z_t)}(f(\theta, z_t))\right)\\
\leq&\frac{2C_0T\epsilon}{\xi},
\end{align*}
where the first inequality follows by the optimality of $\pi_\epsilon^*$, the equality follows by the definition of $\pi_\epsilon$ and the last inequality follows by Assumption~\ref{assump:bounded_norm}\rev{, which implies that $|f(x)|=|\langle f, k(x, \cdot)\rangle|\leq\|f\|\|k(x,\cdot)\|\leq C_0$}. 
\end{proof}

It will be seen that $\epsilon$ plays a key role in trading some regret for strict time-average feasibility.

\subsubsection{Bound Cumulative Regret}

We have the following lemma to bound $\lf{t}(\theta_t, z_t)-\E_{\pi^*_\epsilon(z_t)}(\lf{t}(\theta, z_t))$, which approximates the single-step regret. 
\begin{lemma}
\label{lem:bound_func_change}
$$\lf{t}(\theta_t, z_t)-\E_{\pi^*_\epsilon(z_t)}(\lf{t}(\theta, z_t))\leq2\eta\|C\|^2-\eta\Delta_t,
$$
where $\|C\|^2=\sum_{i=1}^NC_i^2$ {and $\epsilon$ is set to be small enough such that $\epsilon\leq C_i,\forall i\in[N]$}. 
\end{lemma}
\begin{proof}
\begin{align*}
\Delta_t &= V(\lambda_{t+1})-V(\lambda_t)\\
&\leq \lambda_t^T(\lowg{t}(\theta_t, z_t)+\epsilon e) + \frac{1}{2}\|\lowg{t}(\theta_t, z_t)+\epsilon e\|^2\\
&\leq\lambda_t^T\lowg{t}(\theta_t, z_t)+\frac{1}{\eta}\lf{t}(\theta_t, z_t)+\epsilon\lambda_t^Te-\frac{1}{\eta}\lf{t}(\theta_t, z_t) \\
&\qquad + \frac{1}{2}\sum_{i=1}^N(C_i+\epsilon)^2 \\
&\leq\lambda_t^T\E_{\pi^*_\epsilon(z_t)}(\lowg{t}(\theta, z_t))+\frac{1}{\eta}\E_{\pi^*_\epsilon(z_t)}(\lf{t}(\theta, z_t))+\epsilon\lambda_t^Te\\
&\qquad-\frac{1}{\eta}\lf{t}(\theta_t, z_t) + 2\sum_{i=1}^NC_i^2\\
&\leq\frac{1}{\eta}(\E_{\pi^*_\epsilon(z_t)}(\lf{t}(\theta, z_t))-\lf{t}(\theta_t, z_t)) + 2\|C\|^2,
\end{align*}
\rev{where the first inequality follows by the inequality~\eqref{eqn:dual_dec_ineq}, the second inequality follows by adding and subtracting $\frac{1}{\eta}\lf{t}(\theta_t, z_t)$ and the projection operation to $[-C_i, C_i]$ as shown in~\eqref{eqn:lu_defs}, the third inequality follows by the optimality of $\theta_t$ for the primal update problem~\eqref{eqn:primal_udt} and the assumption that $\epsilon\leq C_i$, and the last inequality follows by the feasibility of $\pi_\epsilon^*$ for the problem~\eqref{eqn:rel_epsilon_prob_form}. Rearrangement of the above inequality gives the desired result.} 
\end{proof}


We are then ready to upper bound the cumulative regret. 
\begin{lemma}[Cumulative Regret Bound]
\label{lem:bound_cumu_R}
$$\mathcal{R}_T\leq2\beta_{0,T}^{1/2}\sqrt{4(T+2)\gamma_T}+2\eta T\|C\|^2+\eta V(\lambda_1)+\frac{2C_0T\epsilon}{\xi}.
$$
\end{lemma}
\begin{proof}
\begin{align*}
\mathcal{R}_T \leq& \sum_{t=1}^T\left(f(\theta_t, z_t)-{\E}_{\pi^*(z_t)}(f(\theta, z_t))\right)\\
=&\sum_{t=1}^T\left(f(\theta_t, z_t)-\lf{t}(\theta_t, z_t)\right)\\
&+ \sum_{t=1}^T\left(\lf{t}(\theta_t, z_t)-{\E}_{\pi_\epsilon^*(z_t)}(\lf{t}(\theta, z_t))\right)\\
&+\sum_{t=1}^T\E_{\pi_\epsilon^*(z_t)}(\lf{t}(\theta, z_t)-f(\theta, z_t))\\
&+\sum_{t=1}^T \left(\E_{\pi^*_\epsilon(z_t)}(f(\theta, z_t))-\E_{\pi^*(z_t)}(f(\theta, z_t))\right),
\end{align*}
\rev{where the last inequality follows by that relaxed optimal value ${\E}_{\pi^*(z_t)}(f(\theta, z_t))$ is smaller or equal to the original optimal value, and the equality splits the original term into four terms.
For the first term,   
\begin{align*}
&\sum_{t=1}^T\left(f(\theta_t, z_t)-\lf{t}(\theta_t, z_t)\right)\leq\sum_{t=1}^T2\beta_{0,t}^{1/2}\sigma_t(\theta_t, z_t)\\
\leq&2\beta_{0,T}^{1/2}\sum_{t=1}^T\sigma_t(\theta_t, z_t)\leq2\beta_{0,T}^{1/2}\sqrt{4(T+2)\gamma_T},
\end{align*}
where the first inequality follows by Lem.~\ref{lem:hpb_int}, the second inequality follows by the monotonicity of $\beta_{0,t}^{1/2}$, and the last inequality follows by Lem.~\ref{lem:bound_cumu_sd}.
For the second term, by Lem.~\ref{lem:bound_func_change}, we have,
\begin{align*}
    &\sum_{t=1}^T\left(\lf{t}(\theta_t, z_t)-\E_{\pi_\epsilon^*(z_t)}(\lf{t}(\theta, z_t))\right)\leq\sum_{t=1}^T(2\eta \|C\|^2-\eta\Delta_t)\\
& =2\eta T\|C\|^2+\eta V(\lambda_1)-\eta V(\lambda_{T+1})\leq2\eta T\|C\|^2+\eta V(\lambda_1).
\end{align*}
The third term is non-positive due to Lem.~\ref{lem:hpb_int}. Combining the three bounds and the Lem.~\ref{lem:bound_pert_opt} gives the desired result. 
}
\end{proof}
\subsubsection{Bound Cumulative Violation}
The dual update indicates that violations are reflected in the dual variable. So we first upper bound the dual variable. The idea is to show that whenever the dual variable is very large, it {will be \rev{decreased}}. 
\begin{lemma}
If $V(\lambda_t)\geq\frac{N}{2}\left(\frac{4C_0}{\eta\xi}+\frac{4\|C\|^2}{\xi}\right)^2$ and $0<\epsilon\leq\frac{\xi}{2}$, we have $V(\lambda_{t+1})\leq V(\lambda_t)$.\label{lem:v_func_dec}
\end{lemma}
\begin{proof}
By the primal updating rule,
\begin{align*}    
&\lf{t}(\theta_t, z_{t})+\eta\lambda_t^T\lowg{t}(\theta_t, z_t)\\
\leq &\lf{t}(\bar{\theta}(z_t), z_{t})+\eta\lambda_t^T\lowg{t}(\bar{\theta}(z_t), z_t)\\
\leq &\lf{t}(\bar{\theta}(z_t), z_{t})+\eta\lambda_t^Tg(\bar{\theta}(z_t), z_t)\\
\leq& C_0+\eta(-\xi)\lambda_t^Te,
\end{align*}
\rev{where the first inequality follows by the optimality of $\theta_t$ for the primal update problem, the second inequality follows by that both $\eta$ and $\lambda_t$ are non-negative, and the third inequality follows by Lem.~\ref{lem:hpb_int} and Assumption~\ref{assump:uniform_slater}.}
On the other hand, 
\[
\lf{t}(\theta_t, z_{t})+\eta\lambda_t^T\lowg{t}(\theta_t, z_t)\geq -C_0+\eta\lambda_t^T\lowg{t}(\theta_t, z_t),
\]
\rev{by the Lem.~\ref{lem:hpb_int}.}
Therefore,
\[
C_0+\eta(-\xi)\lambda_t^Te\geq-C_0+\eta\lambda_t^T\lowg{t}(\theta_t, z_t),
\]
which implies
\begin{align*}
\lambda_t^T\lowg{t}(\theta_t, z_t)\leq\frac{2C_0}{\eta}-\xi\lambda_t^Te.
\end{align*}
\rev{So we can get}
\begin{align*}
&V(\lambda_{t+1})-V(\lambda_t)\\\leq& \lambda_t^T(\lowg{t}(\theta_t, z_t)+\epsilon e)+\frac{1}{2}\|\lowg{t}(\theta_t, z_t)+\epsilon e\|^2\\
\leq& \frac{2C_0}{\eta}-\frac{\xi}{2}\lambda_t^Te+\frac{1}{2}\|\lowg{t}(\theta_t, z_t)+\epsilon e\|^2 \\
\leq& \frac{2C_0}{\eta}-\frac{\xi}{2}\sqrt{\frac{\|\lambda_t\|^2}{N}}+2\|C\|^2\leq0,
\end{align*}
\rev{where the first inequality follows by the inequality~\eqref{eqn:dual_dec_ineq}, the second inequality follows by that $\epsilon\leq\frac{\xi}{2}$, the third inequality follows by that $\lambda_t\geq0$ and the Lem.~\ref{lem:hpb_int}, and the last inequality follows by $V(\lambda_t)\geq\frac{N}{2}\left(\frac{4C_0}{\eta\xi}+\frac{4\|C\|^2}{\xi}\right)^2$.}
\end{proof}

Consequently, we have, 
\begin{lemma}
\label{lem:bound_dual}
Let $\lambda_1\leq(\frac{4C_0}{\eta\xi}+\frac{4\|C\|^2}{\xi})e$, we have for any $t$, 
$$
V(\lambda_t)\leq C_V(\eta),
$$
where
$C_V(\eta)=\frac{N}{2}\left(\frac{4C_0}{\eta\xi}+\frac{4\|C\|^2}{\xi}\right)^2+\frac{2C_0}{\eta}+2\|C\|^2.
$
\end{lemma}
\begin{proof}
We use induction. $\lambda_1$ satisfies the conclusion. 


We now discuss conditioned on the value $\lambda_t$. 

\textbf{Case 1}: $V(\lambda_t)\leq\frac{N}{2}\left(\frac{4C_0}{\eta\xi}+\frac{4\|C\|^2}{\xi}\right)^2$,
then $V(\lambda_{t+1})\leq V(\lambda_t)+\frac{1}{\eta}(\E_{\pi^*_\epsilon(z_t)}(\lf{t}(\theta, z_t))-\lf{t}(\theta_t, z_t)) + 2\|C\|^2\leq \frac{N}{2}\left(\frac{4C_0}{\eta\xi}+\frac{4\|C\|^2}{\xi}\right)^2+\frac{2C_0}{\eta}+2\|C\|^2.$

\textbf{Case 2}: $V(\lambda_t)\geq\frac{N}{2}\left(\frac{4C_0}{\eta\xi}+\frac{4\|C\|^2}{\xi}\right)^2$, then $V(\lambda_{t+1})\leq V(\lambda_t)\leq\frac{N}{2}\left(\frac{4C_0}{\eta\xi}+\frac{4\|C\|^2}{\xi}\right)^2+\frac{2C_0}{\eta}+2\|C\|^2$. 
\vspace{0.2cm}

By induction, the conclusion holds for any $t$.
\end{proof}





We now can upper bound the cumulative violation.
\begin{lemma}[Cumulative Violation Bound]
$$
\mathcal{V}_T\leq\left\|\left[\lambda_{T+1}+2\beta_T^g \sqrt{4(T+2)\gamma^g_T}-T\epsilon e\right]^+\right\|\\,
$$
where $\beta_T^g=(\beta_{1,T}, \cdots,\beta_{N,T})$, $\gamma_T^g=(\gamma_{1,T}, \cdots,\gamma_{N,T})$ and multiplication is interpreted elementwise.  
\label{lem:bound_vio}
\end{lemma}
\begin{proof}
By the dual updating rule, we have $\lambda_{t+1}\geq\lambda_t+\lowg{t}(\theta_t, z_t)+\epsilon$. By summing up from $t=1$ to $T$, we get,
\[
\lambda_{T+1}\geq\lambda_1+\sum_{t=1}^T\lowg{t}(\theta_t, z_t)+T\epsilon e.
\]
\rev{Rearranging the above inequality gives,
\begin{equation}
\label{eqn:lowg_cumu_ineq}
 \sum_{t=1}^T\lowg{t}(\theta_t, z_t)\leq\lambda_{T+1}-\lambda_1-T\epsilon e.  
\end{equation}
}
\rev{We thus have,}
\begin{align*}
\sum_{t=1}^Tg(\theta_t, z_t)=& \sum_{t=1}^T\lowg{t}(\theta_t, z_t)+\sum_{t=1}^T(g(\theta_t, z_t)-\lowg{t}(\theta_t, z_t))\\
\leq&\lambda_{T+1}-\lambda_1-T\epsilon e+2\beta_{T}^g\sqrt{4(T+2)\gamma^g_T},\\
\end{align*}
where \rev{the inequality follows by combining the inequality~\eqref{eqn:lowg_cumu_ineq}, the monotonicity of $\beta_T^g$ and Lem.~\ref{lem:bound_cumu_sd}.}.
Therefore,
\begin{align*}
\mathcal{V}_T
=&\left\|{\left[\sum_{t=1}^Tg(\theta_t, z_t)\right]^+}\right\|\\
\leq&\left\|\left[\lambda_{T+1}+2\beta_T^g \sqrt{4(T+2)\gamma^g_T}-T\epsilon e\right]^+\right\|.
\end{align*}
\end{proof}






\subsection{Proof of Thm.~\ref{thm:main_thm}}
\begin{subequations}
\begin{align}
&\lambda_{T+1}+2\beta_T^g \sqrt{4(T+2)\gamma^g_T}\\
\leq&\left(\|\lambda_{T+1}\|+8\|\beta_T^g\|\sqrt{T\|\gamma_T^g\|}\right)e\\
\leq&\Bigg{(}\sqrt{{N}\left(\frac{4C_0}{\eta\xi}+\frac{4\|C\|^2}{\xi}\right)^2+\frac{4C_0}{\eta}+4\|C\|^2}\nonumber\\
&\quad  +8\|\beta_T^g\|\sqrt{T\|\gamma^g_T\|}\Bigg{)}e=T\epsilon e
\end{align}
\end{subequations} 
where the first inequality follows by simple algebraic manipulation and the second inequality follows by Lem.~\ref{lem:bound_dual}. Combining the above inequality and Lem.~\ref{lem:bound_vio} gives
$$
\mathcal{V}_T\finalRev{=}0.
$$
Plugging the values of $\eta, \lambda_1$ and $\epsilon$ into the Lem.~\ref{lem:bound_cumu_R} gives,
$$
\mathcal{R}_T=\mathcal{O}(\sum_{i=0}^N\gamma_{i,T}\sqrt{T}).
$$
}
\end{document}